\documentclass{article}

\usepackage[final,nonatbib]{neurips_2020_ml4ps}

\usepackage{savesym,multirow,url,xspace,graphicx,hyperref,enumitem,color}
\usepackage[dvipsnames]{xcolor}
\usepackage[titletoc]{appendix}
\usepackage{subcaption}
\usepackage{dsfont}
\usepackage{multicol}

\usepackage{wrapfig}

\usepackage{etoolbox}
\usepackage[utf8]{inputenc} 
\usepackage[T1]{fontenc}    
\usepackage{hyperref}       
\usepackage{url}            
\usepackage{booktabs}       
\usepackage{amsfonts}       
\usepackage{nicefrac}       
\usepackage{microtype}      
\usepackage{tikz}
\usepackage{pgfplots}
\usepackage{verbatim}
\usepackage{amsmath,bm}
\usepackage{multirow}
\usepackage{mathtools}
\usepackage{comment}

\newcommand{\argmin}{\operatornamewithlimits{arg\,min}}

\newtoggle{showcomments}
\settoggle{showcomments}{true}

\iftoggle{showcomments}{
\newcommand{\chinmaya}[1]{{\textcolor{purple}{[{\bf Chinmaya:} #1]}}}
\newcommand{\kristin}[1]{{\textcolor{red}{[{\bf Kristin:} #1]}}}
\newcommand{\david}[1]{{\textcolor{NavyBlue}{[{\bf David:} #1]}}}
\newcommand{\yuxin}[1]{{\textcolor{ForestGreen}{[{\bf Yuxin:} #1]}}}
\newcommand{\todo}[1]{{\textcolor{blue}{[{\bf Todo:} #1]}}}
}
{
\newcommand{\chinmaya}[1]{}
\newcommand{\kristin}[1]{}
\newcommand{\david}[1]{}
\newcommand{\yuxin}[1]{}
\newcommand{\todo}[1]{}
}

\makeatletter
\newif\if@restonecol
\makeatother

\usepackage[lined, boxed, ruled, commentsnumbered,noend]{algorithm2e}
\usepackage{amsmath,amssymb,xfrac,amsthm}
\setitemize{noitemsep,topsep=1pt,parsep=1pt,partopsep=1pt, leftmargin=12pt}

\newtheorem*{theorem*}{Theorem}

\makeatletter

\newcommand{\Rmnum}[1]{\expandafter\@slowromancap\romannumeral #1@}
\makeatother
\usepackage{caption}

\newcommand{\figref}[1]{Fig.~\ref{#1}}

\newcommand{\secref}[1]{\S\ref{#1}}

\def \argmin {\mathop{\rm arg\,min}}

\def\BState{\State\hskip-\ALG@thistlm}

\newtoggle{longversion}
\settoggle{longversion}{true}

\title{Towards an Interpretable Data-driven Trigger System for High-throughput Physics Facilities}

\author{Chinmaya Mahesh\textsuperscript{\textdagger} \quad Kristin Dona\textsuperscript{*} \quad David W. Miller\textsuperscript{*} \quad Yuxin Chen\textsuperscript{*}\\
\textsuperscript{\textdagger} 
University of Illinois Urbana-Champaign, \texttt{cmahesh2@illinois.edu}\\
\textsuperscript{*}University of Chicago, 
\texttt{\{kdona, davemilr, chenyuxin\}@uchicago.edu} \\
}

\begin{document}

\maketitle

\begin{abstract}
    Data-intensive science is increasingly reliant on real-time processing capabilities and machine learning workflows, in order to filter and analyze the extreme volumes of data being collected. This is especially true at the energy and intensity frontiers of particle physics where bandwidths of raw data can exceed 100~Tb/s of heterogeneous, high-dimensional data sourced from hundreds of millions of individual sensors. In this paper, we introduce a new data-driven approach for designing and optimizing high-throughput data filtering and trigger systems such as those in use at physics facilities like the Large Hadron Collider (LHC). 
    Concretely, our goal is to design a data-driven filtering system with a minimal run-time cost for determining which data event to keep, while preserving (and potentially improving upon) the distribution of the output as generated by the hand-designed trigger system. 
    We introduce key insights from interpretable predictive modeling and cost-sensitive learning in order to account for non-local inefficiencies in the current paradigm and construct a cost-effective data filtering and trigger model that does not compromise physics coverage. 
    
\end{abstract}
\section{Introduction and Background}

Data filtering algorithms --- or \textit{trigger algorithms}~\cite{ATLASTDAQTDR2003,CMSTDAQTDR2000,CMSTDAQTDR2002} --- targeted at discovery science (e.g. identification of a data event containing evidence of dark matter produced at the Large Hadron Collider), must operate at the level of 1 part in $10^5$ due to numerous bandwidth, computation, and storage-related constraints. Once executed, these algorithms often drive the data curation process, funneling event records with certain features into categories that are predefined based on the labels extracted by those algorithms. 
Although discovery science has been based on this approach for decades, the design of these systems relies heavily on excellent prior knowledge of the feature space being probed, detailed simulations of hand-designed feature extraction algorithms, and heavily modularized approaches to optimization. Consequently, significant inefficiencies exist in these systems, including both redundant feature labeling schemes and cost-ineffective algorithm execution. 

\begin{figure}
  \centering
  \includegraphics[scale=0.44]{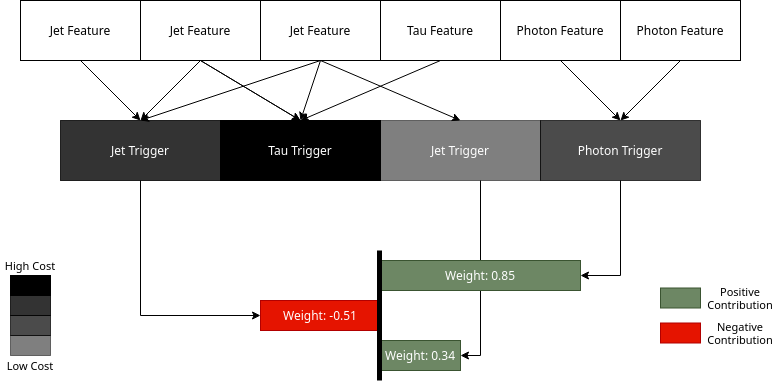}
  \caption{An example cost-effective explanation of an event. In this case, the Tau Trigger has the highest cost and thus the weight associated with the Tau Trigger was driven to 0. The remaining weights result in the final cost-effective explanation of the event, with the weights with the highest absolute value being the most important.}\label{fig:explain}
\end{figure}

In this paper, we describe the design and initial results of the construction of a data-driven approach to refine the trigger and data filtering algorithms at future physics facilities. To accomplish such a daunting task, it is essential that the \textit{explainability} of such an approach first be established. Whereas in other contexts, the explainability of a classifier decision, for example, can be related to expert variables that describe the physics of a process~\cite{Guest_2016,faucett2020mapping,collado2020learning,agarwal2020explainable}, we use the relationship to a hand-designed set of trigger decisions in the approach described here. As illustrated in Figure~\ref{fig:explain}, we may therefore construct a cost model by which the data filtering and curation process can be quantifiably assessed. This approach allows for \textit{pruning} of costly but ineffective individual algorithm labels. In addition, progress towards a fully functional automated data filtering and processing system must start by constructing a learning-based filter system that is able to reproduce the current, hand-designed, trigger ``menus.'' More importantly, though, in order to be useful, this new system must be able to \textit{explain} its selections in the context of both low-level features (e.g. sensor signals) and high-level features (e.g. physics object multiplicities and kinematics). This is the first step towards the ultimate goal of constructing an active continuous learning model that is able to update itself and provide \textit{explanations} for those updates.  A distinguishing aspect of this research effort is its focus on model \textit{interpretability}: Instead of a closed-box model that is capable of recovering the original data distribution, we aim to design an \textit{open-box} predictive model, which, for any given input, not only outputs a decision (e.g., ``keep this data point''), but also explains why we should do so, by associating the decision with the existing rules in the hand-designed trigger menu.

In addition to the development of an open-box predictive model, we also aim to minimize the latency of the trigger system. 
Given an incoming data event, each trigger algorithm incurs a latency at runtime -- assuming that algorithms are run in parallel, the latency of the trigger system depends on the worst-case running time of all trigger algorithms. Thus, for each data event, finding the most efficient set of trigger algorithms at run time is crucial for a real-time trigger system. 

To address the real-time data processing challenge, we investigate the following combinatorial optimization problem: given a ground set of candidate trigger algorithms from the existing trigger menu and a latency cost for each trigger algorithm, we seek an optimal subset of trigger algorithms for each incoming data event, such that the selected algorithms can jointly make the correct filtering decision with the \emph{minimal latency cost}. We then use the solution of this optimization problem as the explanation of our open-box predictive model, which in turn will be used to optimize the latency of the existing trigger system. Here, we highlight a significant challenge during this process: Since the effectiveness of any subset of trigger algorithm is often unknown \emph{a priori} and needs to be learned from data, this task essentially amounts to building a data-driven, cost-sensitive explainable model---an emerging topic at the frontier of machine learning and operations research.
\section{Related Work}\label{sec:related}

\paragraph{Interpretable models}
Interpretable predictive modeling is useful for very wide application domains where black-box models are not preferred (e.g., medical / biological research, emergency response planning, etc). In these scenarios, decisions are critical and often have serious consequences, and domain experts hence would like to know how theses predictions are made. Depending on different measures of \emph{interpretability}, existing methods can be categorized into three types: \emph{sparse models}, which use a small number of features / parameters, such as sparse linear classifiers \cite{chang2009reading, ustun2014methods}, (low-rank) latent factor models, etc; \emph{discretization methods}, which split up a problem into several (independent) sub-problems, such as decision trees and decision lists \cite{letham2015interpretable}; and \emph{prototype-based classifiers}, which use examples and basic features instead of formal models and constructed features, such as nearest neighbors \cite{cover1967nearest}. Recently developments in interpretable models, e.g., Local interpretable model-agnostic explanations (LIME) \cite{ribeiro2016should}, have proposed to use local surrogate models to explain individual predictions of black box machine learning models, where each local surrogate model is interpretable by itself. Our work extends the LIME framework and can be viewed as a sparsity-based locally interpretable model, where we seek a minimal-cost explanation for the LHC trigger outputs.
\paragraph{Resource-efficient machine learning}
In many data-driven decision making tasks such as active learning \cite{tong2001support,tong2001image,settles2012active,settles2011closing}, recommender systems \cite{lattimore2020bandit,chen17onlinevoi}, experimental design \cite{chaloner1995bayesian} and surveillance \cite{loy2012stream,chu2011unbiased, bouguelia2013stream,chen14active}, learning and prediction are often under stringent resource constraints---As an example, features are associated by costs (e.g. monetary cost for conducting a medical tests, computational cost for extracting a feature) both in training \cite{liu2004selective,raghavan2006active} and test time \cite{gottlieb2012attention,early2016cost}. In connection to our work, a data-driven trigger system must maintain a throughput constraint to keep pace with arriving traffic of the collision event. A common challenge among these applications is the need to properly trade-off prediction accuracy against the prediction cost \cite{golovin2011adaptive,crammer2004online,kumar2017resource}. Our work aligns with contemporary models that demand cost-effectiveness at run time.

\section{Automated Trigger Menu Design}

We are given a dataset $X \in \mathbb{R}^{n \times p}$ of physical collision data (where each of the $n$ records are known as \textit{events}), a set of labels $T$ (known as \textit{triggers}), and an outcome matrix $y = \{0,1\}^{n \times |T|}$ which denotes the triggers each event satisfies. Let each event in $X$ be described by numerical values of a set of $p$ features $F = \{f_1, f_2, ..., f_p\}$. We are also given a mapping $c: F \rightarrow \mathbb{R}^+$, denoting the \textit{cost function}, that is, $c(f_i)$ is the cost of using feature $f_i$ to predict the outcome of an event. The goal is to identify the most \textit{cost-efficient} subset $F' \subset F$ that enables us to maximize coverage of $X$ in the trained model while using only the features in $F'$ to make predictions.

\subsection{LIME with Elastic Net}
We utilize the LIME framework \cite{ribeiro2016should}, which creates an interpretable prediction for an event $x$ by training a sparse model with a dataset of perturbations of $x$. Using the trained weight vector $\hat{\beta}$ of this sparse model we are able to understand the importance of each feature in the final prediction, as a higher value of $|\hat{\beta}_i|$ indicates higher influence of feature $f_i$ in the final prediction. In order to obtain an interpretable yet accurate prediction of the outcome of an event, we consider to apply a sparsity regularizer to the local linear model being trained. As discussed in \secref{sec:related}, interpretability is achieved because the non-zero entries of a sparse weight vector of a linear model can be inspected manually. Note that sparse models might be far less accurate, and how to strike a good balance between performance and interpretability remains an open problem in many practical applications. In this work, we adopt \textit{elastic net} \cite{zou2005regularization} as a general formulation that trades off model interpretability (sparsity) and accuracy. Formally,
given a dataset $X \in \mathbb{R}^{n \times p}$, an outcome vector $y \in \mathbb{R}^n$, and a tunable parameter $\lambda > 0$, the final $\hat{\beta}$ obtained can be defined as $$
\hat{\beta} = \argmin_\beta \left( |y - X\beta|^2 + (1 - \alpha)\lambda \sum_{i=1}^{p} |\beta_i| + \alpha\lambda \sum_{i=1}^{p} |\beta_i|^2 \right).
$$ 
The LASSO and ridge regressions are special cases of the elastic net where $\alpha = 0$ and $\alpha = 1$ respectively.

\subsection{Cost-Effective Elastic Net}
To obtain a value of $\hat{\beta}$ which is both sparse and cost efficient, we propose adding a coefficient of $c(f_i)$, which is the cost of feature $i$, to each respective term $|\beta_i|$ and $|\beta_i|^2$ in the elastic net penalty:
$$
\hat{\beta} = \argmin_\beta \left( |y - X\beta|^2 + (1 - \alpha)\lambda \sum_{i=1}^{p} |\beta_i|\cdot c(f_i) + \alpha\lambda \sum_{i=1}^{p} |\beta_i|^2 \cdot c(f_i) \right)
$$
As we have the restriction $\forall f_i \in F, c(f_i) \in \mathbb{R}^+$, the above modification to the elastic net (henceforth referred to as the \textit{cost-effective elastic net}), maintains all the convexity properties of the elastic net. Thus by using LIME with the cost-effective elastic net as the sparse model, we have modified LIME to create interpretable explanations which additionally have the property of being cost-effective.
\subsection{LIME with Submodular Pick}
With the above approach, we are able to obtain information about the features which are the most important while being cost-effective. However, a larger model-wide explanation to analyse the scenarios in which certain features are determined be both important to the prediction and cost-effective is desired. To achieve this, a modified version of SP-LIME \cite{ribeiro2016should} is used. While calculating the set of representative explanations for the whole model, SP-LIME creates an importance vector $I$ such that if $I_i > I_j$ for any $i$ and $j$, then feature $i$ is considered to be more important than feature $j$. This gives us a total ordering of $F$ which enables us to select an optimal subset of $F$ containing the most important and cost-effective features for any arbitrary cutoff. We call this method of using a modified SP-LIME with a cost-effective elastic net CE-LIME.
\section{Experiments}

\paragraph{Toy Dataset in a Controlled Environment}
In order to empirically test the effectiveness of the cost-effective elastic net, 100 trials were run on randomly generated classification problems created using the \texttt{make\_classification} feature of scikit-learn \cite{scikit-learn}. In each trial, a matrix of predictor variables $X \in \mathbb{R}^{1000 \times 80}$ and an outcome vector $y \in \{0,1\}^{1000}$ were created such that 20 of the features were indicative, 20 were linear combinations of the indicative features, and 20 were duplicated features sampled from the previous two groups of features, in order to create the usual scenario of a high dimensional dataset with several redundant or correlated features. Furthermore, a cost function $c$ was created with a uniformly randomized distribution in the interval $[0,10]$. For any event $x_{\text{new}}$ to be predicted, initially the value of every feature $f_i$ to be used by the classifier is set to the mean of the feature's value in $X$. In this way, a higher accuracy level can be obtained by iteratively adding the value of more features in accordance to the total feature order $I$.

\begin{figure*}[t!]
  \centering
  \begin{subfigure}[b]{.45\textwidth}
    \centering
    {
      \includegraphics[trim={0pt 0pt 0pt 0pt}, width=\textwidth]{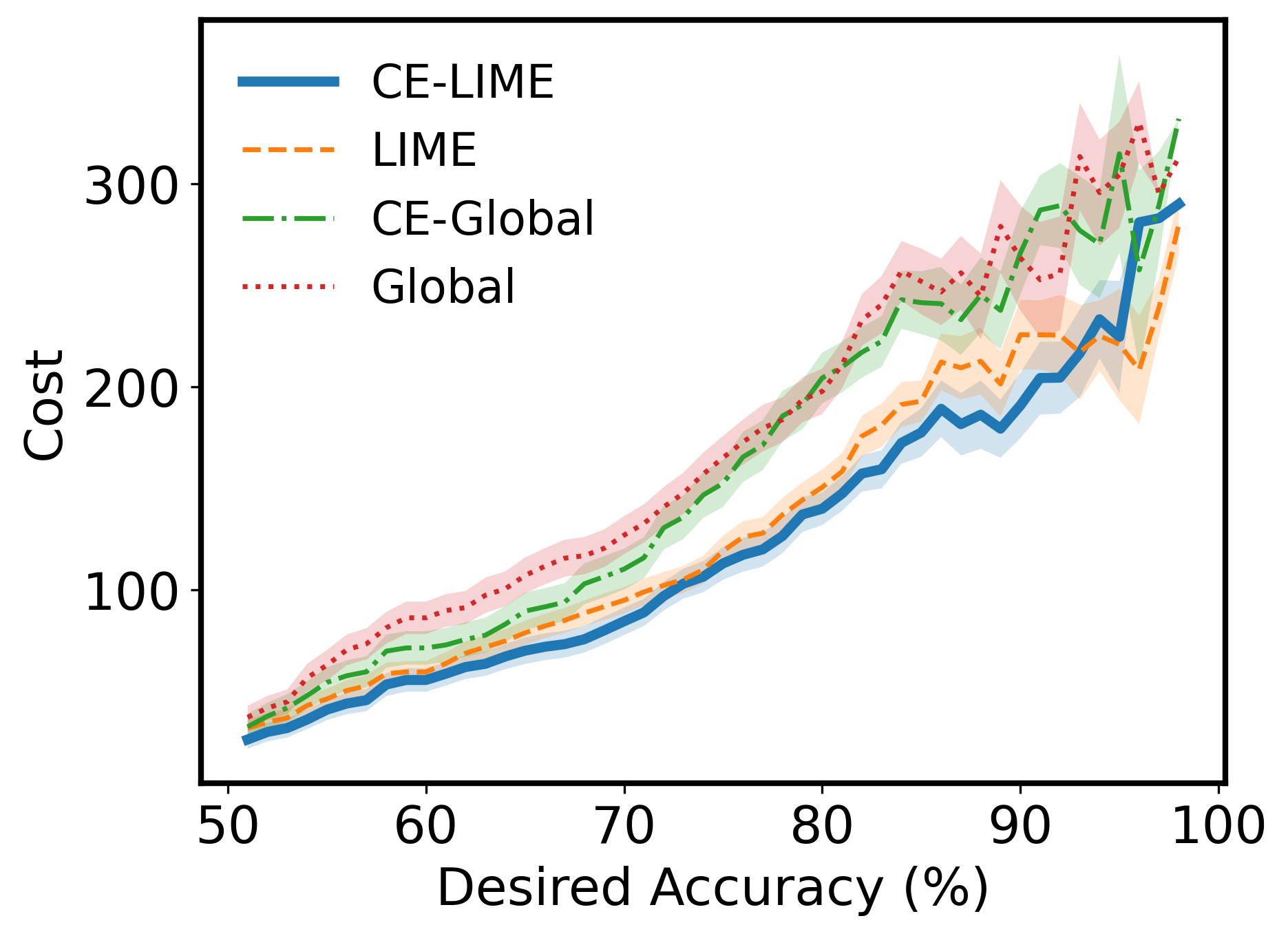}
      \caption{Cost vs Performance, Toy Dataset}
      \label{fig:cvpt}
    }
  \end{subfigure}
  \quad
  \begin{subfigure}[b]{.43\textwidth}
    \centering
    {
      \includegraphics[trim={0pt 0pt 0pt 0pt}, width=\textwidth]{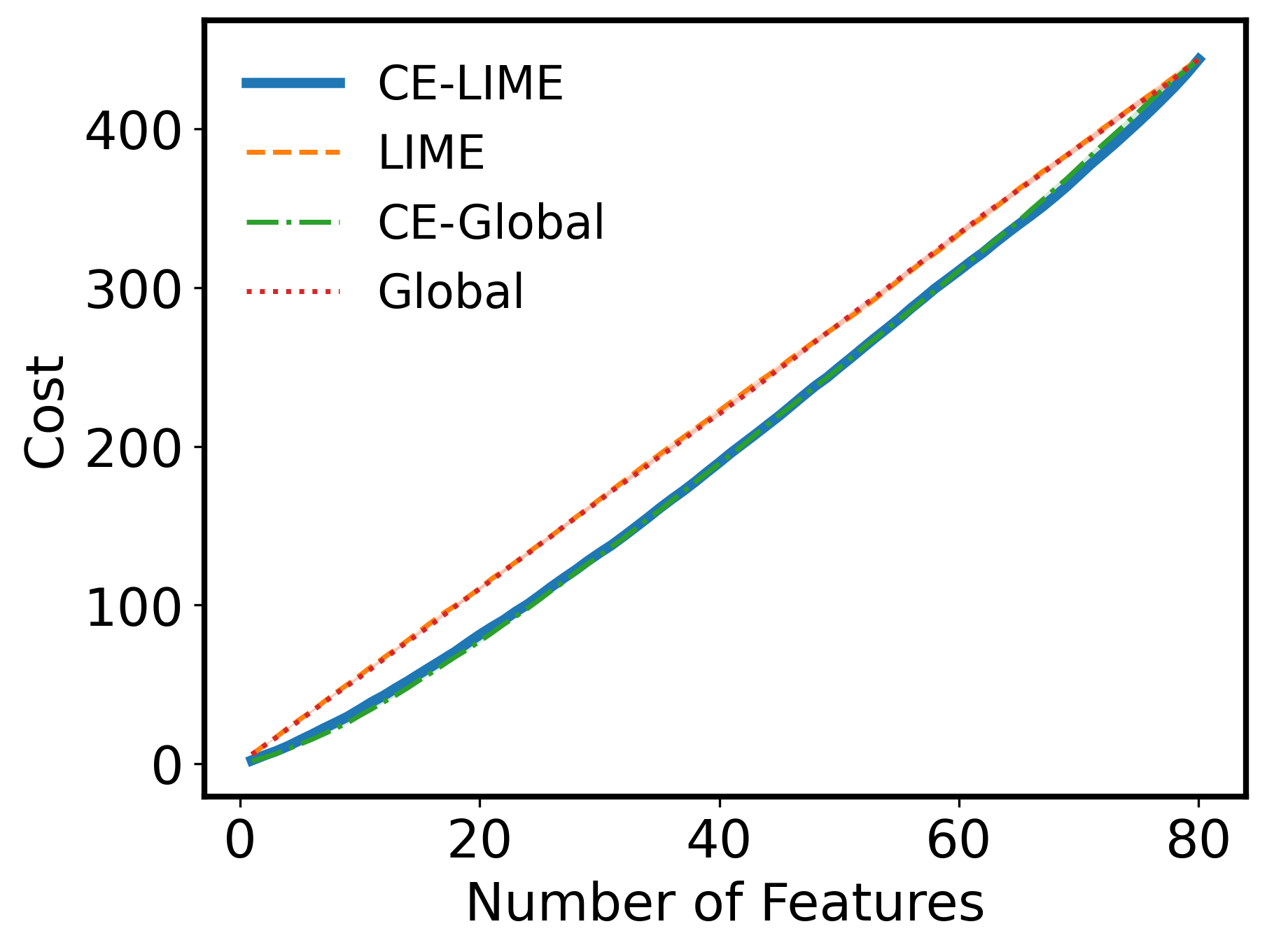}
      \caption{Cost vs \# Used Features, Toy Dataset}
      \label{fig:cvut}
    }
  \end{subfigure}
  \begin{subfigure}[b]{.43\textwidth}
    \centering
    {
      \includegraphics[trim={0pt 0pt 0pt 0pt}, width=\textwidth]{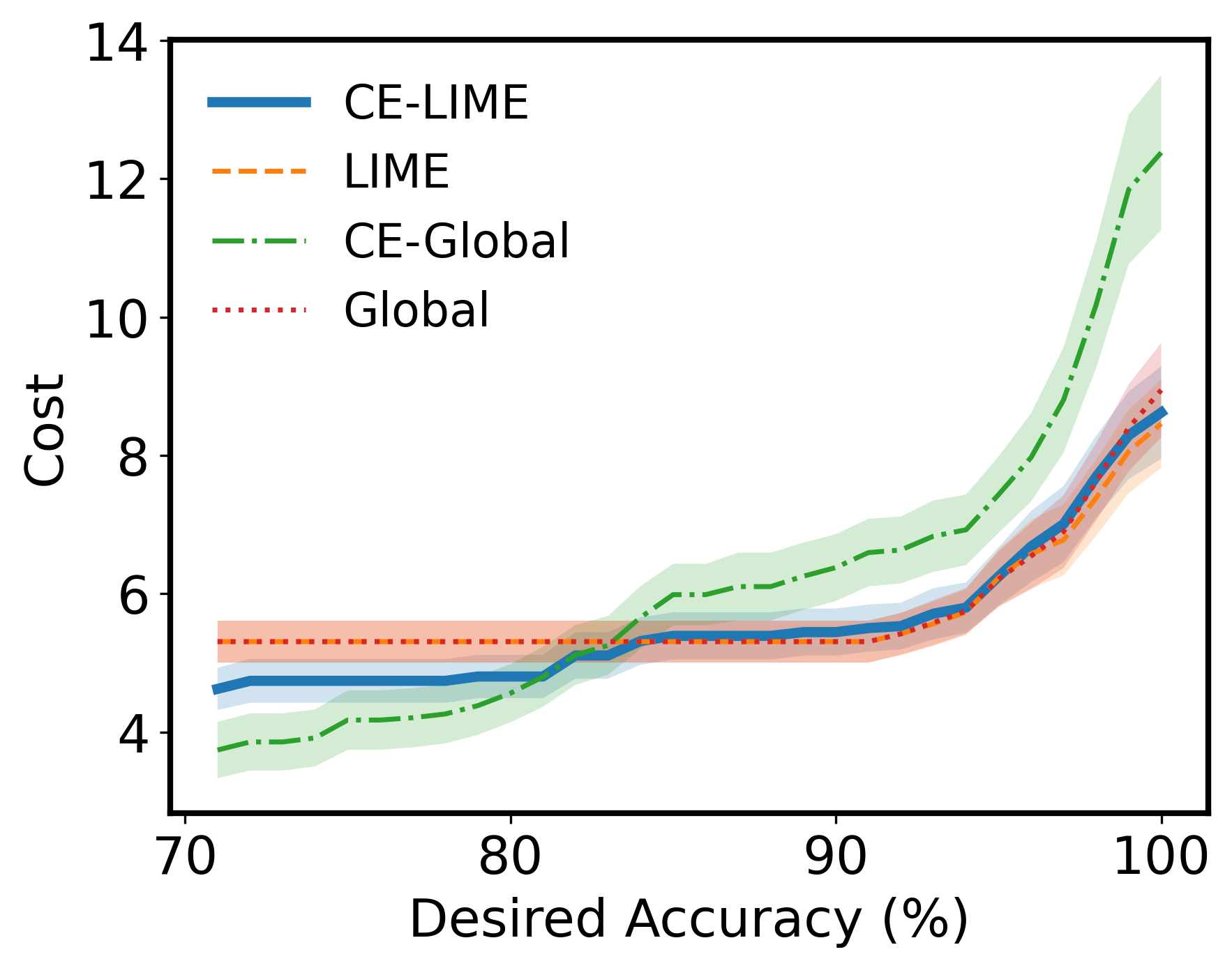}
      \caption{Cost vs Performance, CMS Open Data}
      \label{fig:cvpe}
    }
  \end{subfigure}
  \quad
  \begin{subfigure}[b]{.43\textwidth}
    \centering
    {
      \includegraphics[trim={0pt 0pt 0pt 0pt}, width=\textwidth]{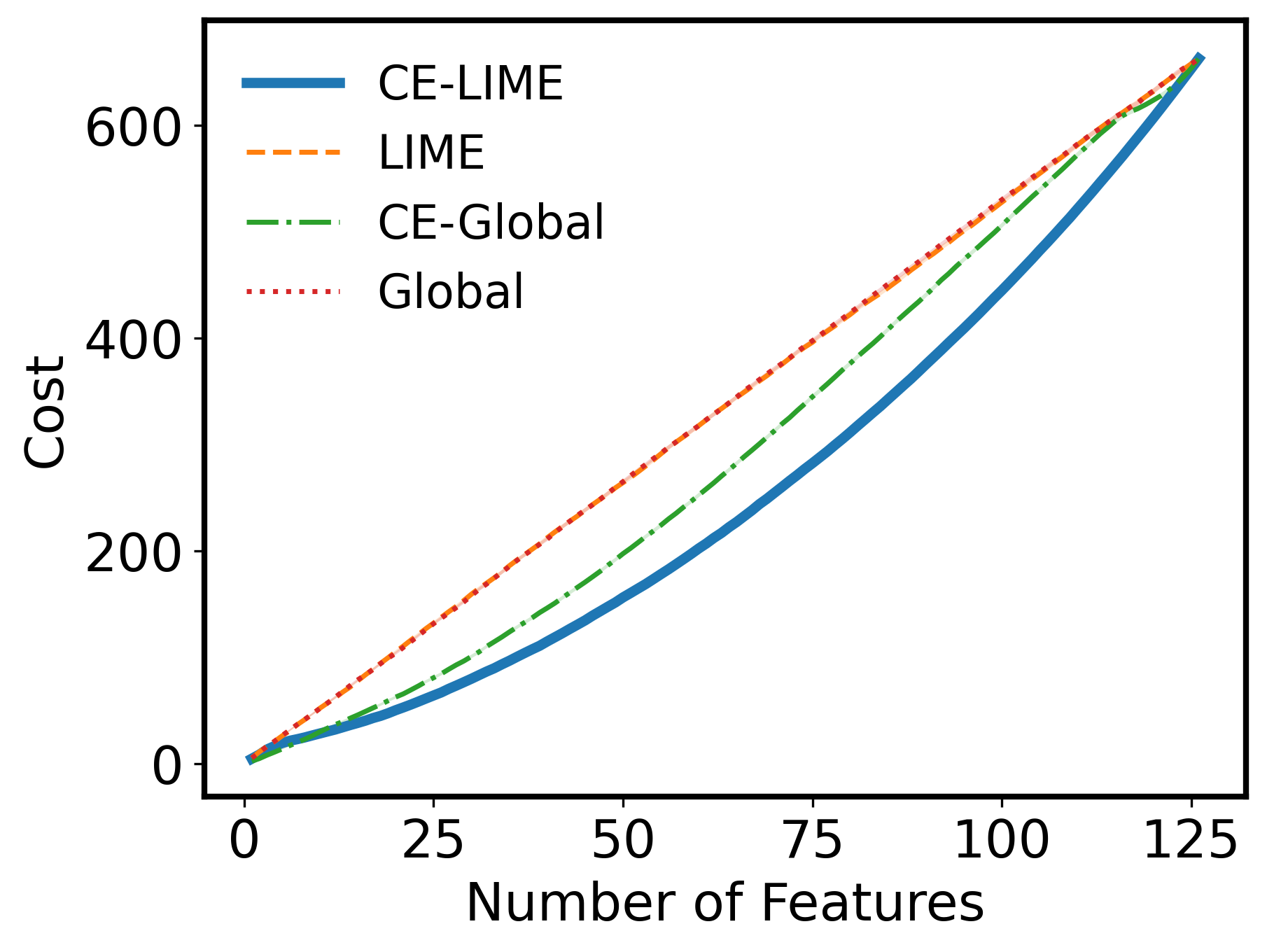}
      \caption{Cost vs \# Used Features, CMS Open Data}
      \label{fig:cvue}
    }
  \end{subfigure}
    \caption{\small Experimental results on synthetic data (top) and CMS Open Data (bottom). We compare our proposed algorithm cost-effective LIME (CE-LIME) with three baseline approaches, namely LIME with submodular pick (LIME), a sparse regressor via elastic net regression (Global), and a cost-effective elastic net (CE-Global). Error bars represent standard error over all trials.}
  \label{fig:exp:setcover:mse}
  \vspace{-3mm}
\end{figure*}

In \figref{fig:cvpt}, we see that CE-LIME can achieve the same accuracy level as LIME with submodular pick (LIME), a sparse regressor via elastic net regression (Global), and a cost-effective elastic net (CE-Global) with an overall lower cost. In \figref{fig:cvut}, we observe a dip of the required cost at certain accuracy levels due to the decrease in probability of classifiers obtaining high accuracy levels. This leads to only the surviving models being counted, which tend to have lower overall cost requirements due to the simpler nature of their respective classification problems.


\vspace{-3mm}
\begin{figure}[!h]
  \centering
  \begin{subfigure}[b]{.49\textwidth}
    \centering
    {
      \includegraphics[trim={0pt 0pt 0pt 0pt}, width=\textwidth]{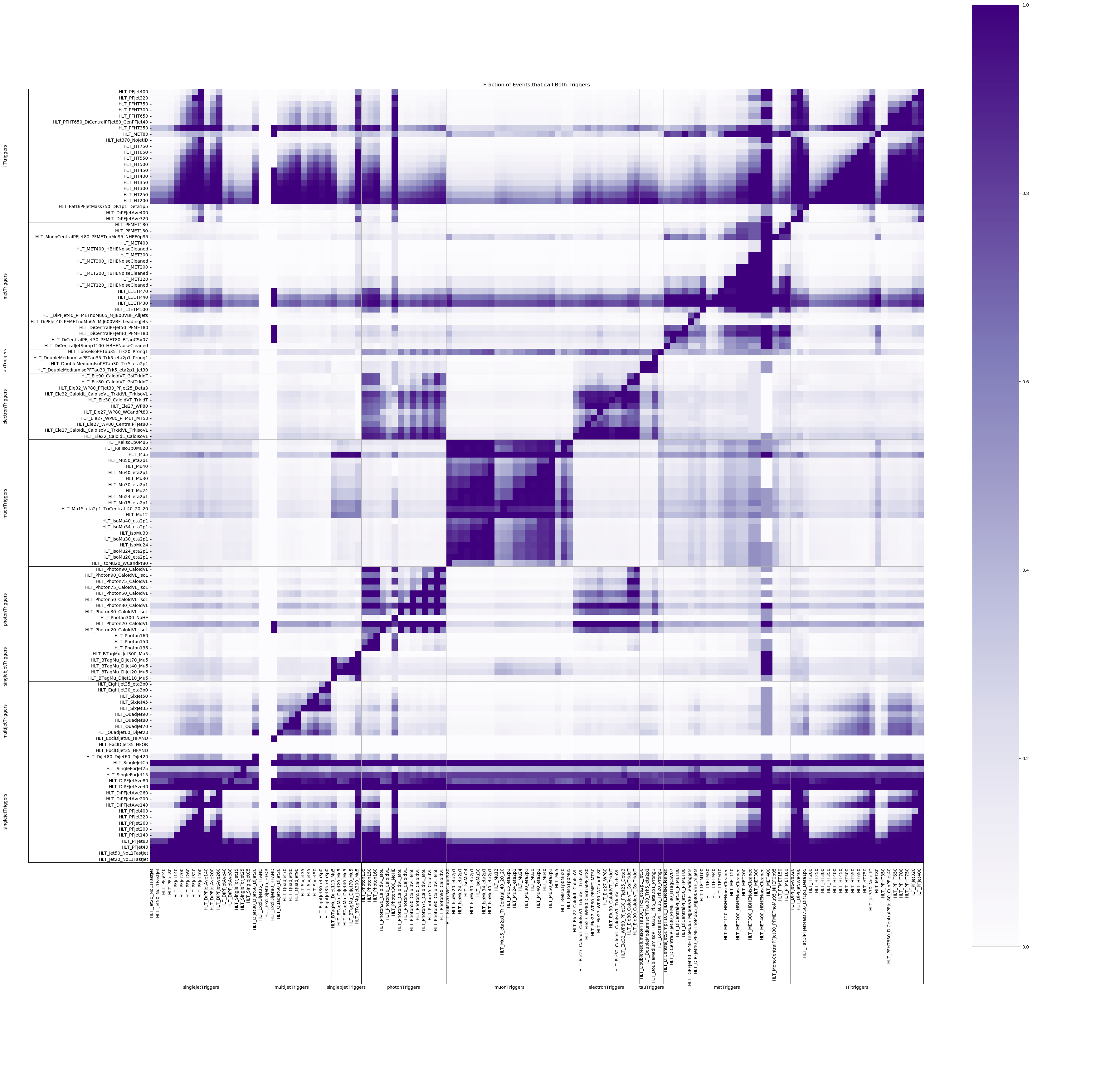}
      \caption{All trigger labels}
      \label{fig:triggeroverlap-all}
    }
  \end{subfigure}
  ~
  \begin{subfigure}[b]{.49\textwidth}
    \centering
    {
      \includegraphics[trim={0pt 0pt 0pt 0pt}, width=\textwidth]{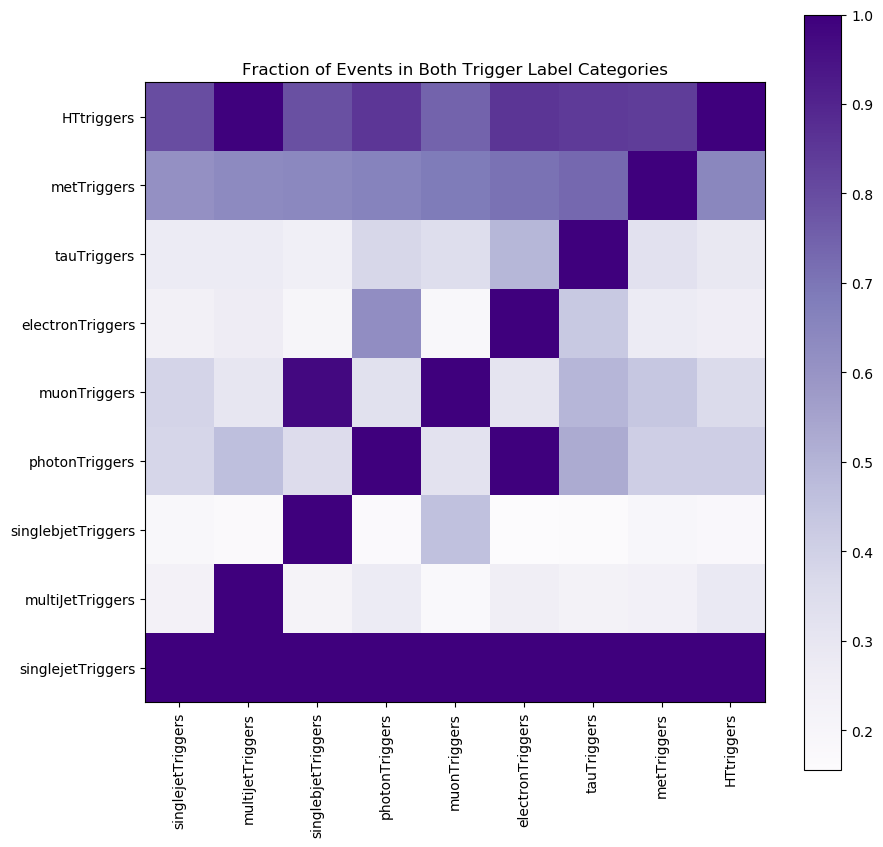}
      \caption{Trigger label categories}
      \label{fig:triggeroverlap-categories}
    }
  \end{subfigure}
  \caption{Trigger label overlap amongst (\subref{fig:triggeroverlap-all}) all labels and (\subref{fig:triggeroverlap-categories}) groups of related triggers separated into categories.}
  \vspace{-3mm}
  \label{fig:triggeroverlap}
\end{figure}

\paragraph{Case Study on CMS Open Data} 
To demonstrate the potential value of these optimization methods at the LHC, we use publicly available datasets from the CMS Experiment at the LHC, available in the public CERN Open Data Portal~\cite{MITOpenData}. The Portal represents an exciting opportunity for the general public to access LHC data for studies such as the trigger optimization presented in this paper.
In particular, we use a simulated dataset of $t\bar{t}$ events \cite{CMSOpenData}, generated for collisions at a center-of-mass energy of 8 TeV and reconstructed to provide information on the physics objects such as electrons, photons, muons, taus, and jets, as well as their respective masses, transverse momenta, and pseudorapidity. The events also include Boolean values for over 400 different triggers that we have analyzed using the physics information and trigger decisions.

Similar to the experimental setup of the toy dataset, 10 trials were run on 9 different triggers with randomized cost of features in every trial, with the costs being uniformly distributed in $[0,10]$. Figure~\ref{fig:triggeroverlap} is generated with the CMS Open Data and shows the fractional overlap between features which share trigger labels and trigger label categories. The large fractional overlap emphasises the potential for these algorithms to be optimized. \figref{fig:cvpe} shows that the required cost to achieve any accuracy level is the lowest for CE-LIME for most cases, and Figure \ref{fig:cvue} shows that the cost of using a certain number of features in the order given by CE-LIME is competitive with the globally trained cost-effective elastic net (CE-GLOBAL).


\section*{Acknowledgment}
The authors thank Dr. Cecilia Tosciri for the helpful discussion. The authors also thank the CMS Open Data Workshop for Theorists for assistance with accessing the CMS public data\footnote{\href{https://indico.cern.ch/event/882586/}{https://indico.cern.ch/event/882586/}}. This work was done when Chinmaya Mahesh was at the Center for Data and Computing (CDAC) Data \& Computing Summer Lab at UChicago. The project was supported by a CDAC Discovery Grant. 

\section*{Broader Impact}


The broader impact of this work derives both from the nexus of novel ideas and the application to which we've applied those ideas. Interpretable predictive modeling combined with the approaches to cost-sensitive learning represent a new direction that aims to introduce a quantifiable metric by which to optimize black-box systems, and then to provide a mechanism for ``opening'' that box. In addition, the novel application of these ideas to the challenging real-time data filtering and processing systems that are in use at large-scale physics facilities offer the potential for a paradigm shift in the operational models, and perhaps scientific capabilities, of those facilities. Future work aims to explore the use of active learning models that capitalize on the interpretability and cost-sensitive optimization discussed in this work to provide for a continuous learning model that is able to ``explain'' what it has learned in terms of both cost and performance.



\bibliographystyle{unsrt}
\bibliography{reference,references,bibliography}


\iftoggle{longversion}{
 \newpage
\onecolumn
\appendix
{\allowdisplaybreaks
}
}
{}
\end{document}